\documentclass{mva_style}
\usepackage{graphicx}
\usepackage{amssymb}
\usepackage{amsmath}
\usepackage{mathtools}
\usepackage{caption}
\usepackage{subcaption}
\finalcopy

\begin{document}
\title{Measuring robustness of Visual SLAM}

	\author
	{
  		David Prokhorov\\
  		{\tt ksetherg@gmail.com}\\
  		\and
  		Dmitry Zhukov\\
  		{\tt d.zhukov@samsung.com}\\
  		\and
    	Olga Barinova\\
  		{\tt o.barinova@samsung.com}\\
  		\and
  		Anna Vorontsova\\
  		{\tt a.vorontsova@samsung.com}\\
  		\and
  		Anton Konushin\\
  		{\tt a.konushin@samsung.com}\\
  		\and
  		\\
  		{\tt }\\
  		\and
  		Samsung AI Center, Moscow \\
  		{\tt }\\
	}

\maketitle

\section*{\centering Abstract}
\textit
{
Simultaneous localisation and mapping (SLAM) is an essential component of robotic systems. In this work we perform a feasibility study of RGB-D SLAM for the task of indoor robot navigation. Recent visual SLAM methods, e.g. ORBSLAM2 \cite{mur2017orb}, demonstrate really impressive accuracy, but the experiments in the papers are usually conducted on just a few sequences, that makes it difficult to reason about the robustness of the methods. Another problem is that all available RGB-D datasets contain the trajectories with very complex camera motions. In this work we extensively evaluate ORBSLAM2 to better understand the state-of-the-art. First, we conduct experiments on the popular publicly available datasets for RGB-D SLAM across the conventional metrics. We perform statistical analysis of the results and find correlations between the metrics and the attributes of the trajectories. Then, we introduce a new large and diverse HomeRobot dataset where we model the motions of a simple home robot. Our dataset is created using physically-based rendering with realistic lighting and contains the scenes composed by human designers. It includes thousands of sequences, that is two orders of magnitude greater than in previous works. We find that while in many cases the accuracy of SLAM is very good, the robustness is still an issue.
}
\section{Introduction}
Simultaneous localization and mapping (SLAM) is the key component of many robotics, augmented reality and 3D scanning systems. The goal of visual SLAM is to estimate the camera poses and simultaneously reconstruct the environment using the images from a camera. For the case of indoor navigation RGB-D SLAM methods that work with 3D sensors \cite{kerl2015dense, whelan2016elasticfusion, mur2017orb} are most suitable since most robotic platforms are equipped with RGB-D cameras. We have tried different implementations of the recent SLAM methods and have chosen state-of-the-art ORBSLAM2 \cite{mur2017orb} due to it's performance and stability of the implementation \footnote{We have tried out all the methods from SLAMBench \cite{bodin2018slambench2} in our experiments, but unfortunately the provided implementations were unstable, that did not allow us to perform large-scale experiments. In the end, we switched to the original implementation of ORBSLAM2 because it could better reproduce the results reported in the original paper \cite{mur2017orb}}.

Benchmarking of SLAM systems for indoor applications requires a dataset of image sequences with ground truth camera poses. Usually, motion capture is used to obtain camera poses for real-world image sequences \cite{sturm2012benchmark}. Another option is to use photo-realistic 3D rendering \cite{handa2014benchmark, li2018interiornet} that provides accurate camera poses. The most popular accuracy metrics for SLAM are Absolute Trajectory Error (ATE) and Relative Position Error (RPE) that measure the quality of localization.

Most of the research papers on RGB-D SLAM are compared by selected sequences from real world TUM RGB-D \cite{sturm2012benchmark} or rendered ICL-NUIM \cite{handa2014benchmark} benchmarks and demonstrate really impressive accuracy in these experiments. However, only a small number of sequences is usually used in the experiments. 

As far as we know this paper is the first effort to perform a large-scale statistical analysis of the accuracy of visual SLAM methods. Our contributions are two-fold. First, we evaluate ORBSLAM2 across all open benchmarks and analyze the correlations between the attributes of the sequences and the accuracy of SLAM. As a byproduct, we perform a degrading experiment with turning off the loop closure component of ORBSLAM2 and surprisingly find that loop closure does not significantly influence the robustness of the system. Second, we introduce a new large and diverse HomeRobot dataset of indoor sequences consisting of over a thousand trajectories. It targets indoor robotics applications and is much larger than ICL-NUIM and the open part of InteriorNet. We evaluate ORBSLAM2 on different modifications of the new dataset and extensively analyze the influence of the speed of agent on the accuracy of localization. 

The structure of the paper is: in Section $2$, we overview standard metrics for measuring the quality of SLAM algorithms. In Section $3$, we describe the datasets that we used. In Section $4$, we describe our experiments and present the analysis of obtained results. In Section $5$, we discuss the results and conclude the paper.
\begin{figure*}[t]
  \centering
	\begin{subfigure}{0.32\textwidth}
		\includegraphics[width=\textwidth]{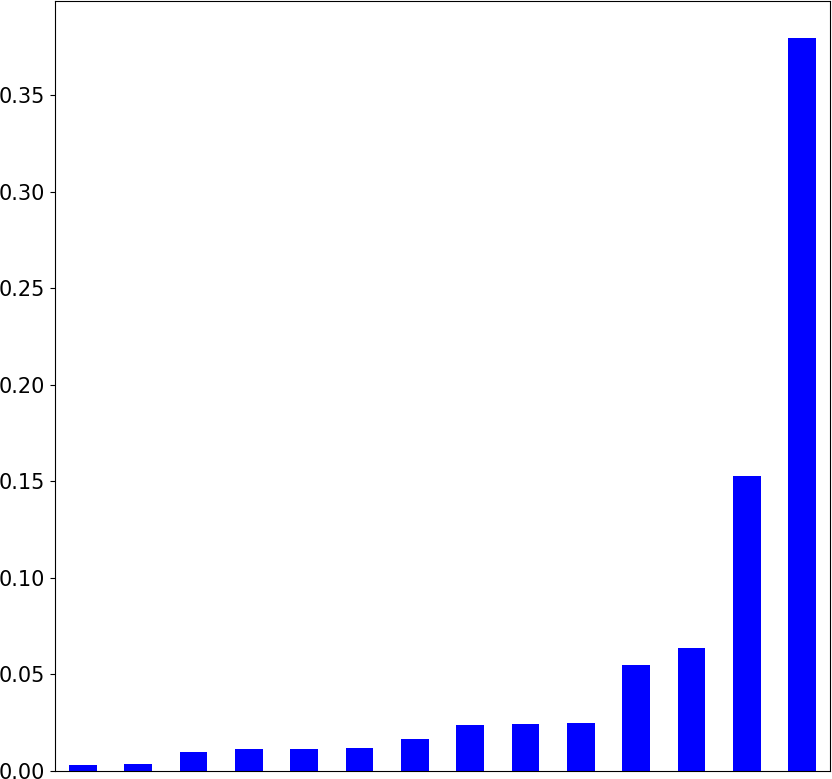}
		\caption{ATE}
		\label{bar-tum-ate}
	\end{subfigure}
	\begin{subfigure}{0.32\textwidth}
    	\includegraphics[width=\textwidth]{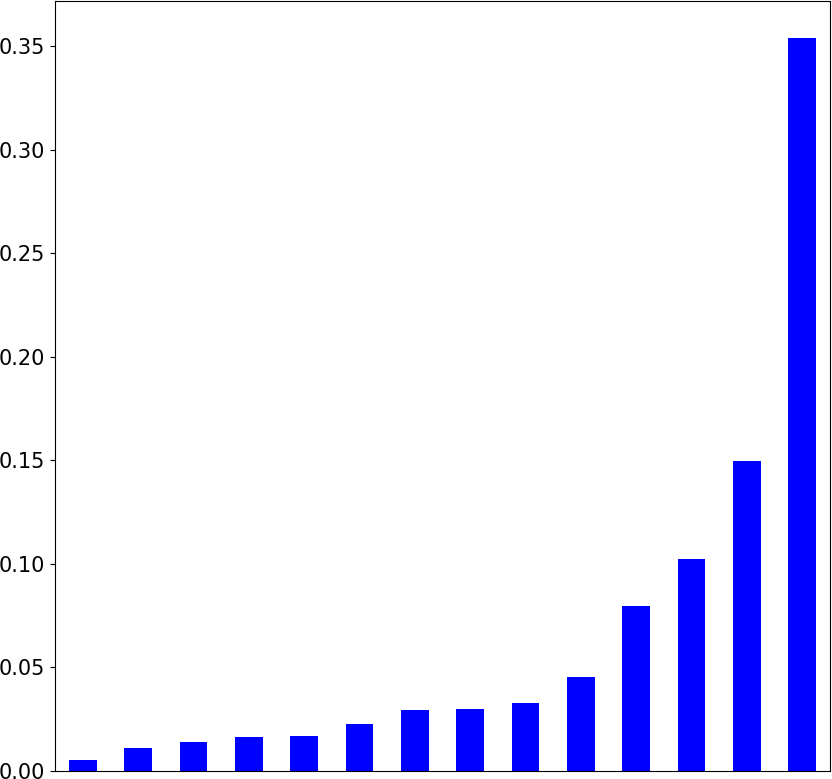}
    	\caption{RPE translation}
    	\label{bar-tum-rpe-trans}
    \end{subfigure}
    \begin{subfigure}{0.32\textwidth}
    	\includegraphics[width=\textwidth]{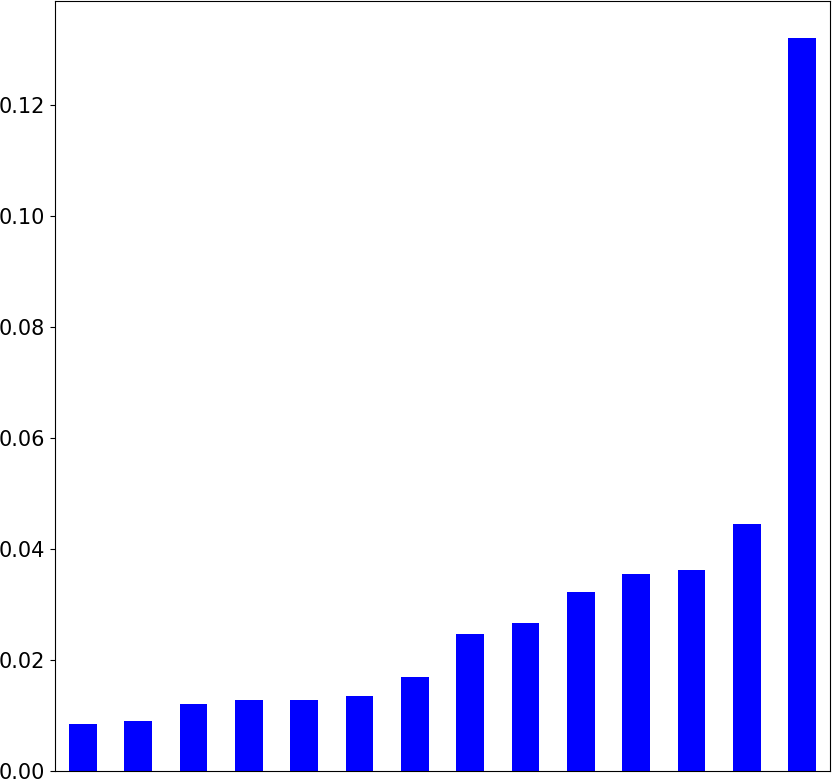}
    	\caption{RPE rotation}
    	\label{bar-tum-rpe-rot}
    \end{subfigure}
	\label{bar-icl-tum}
	\newline
 \centering
  \begin{subfigure}{0.32\textwidth}
    \includegraphics[width=\textwidth]{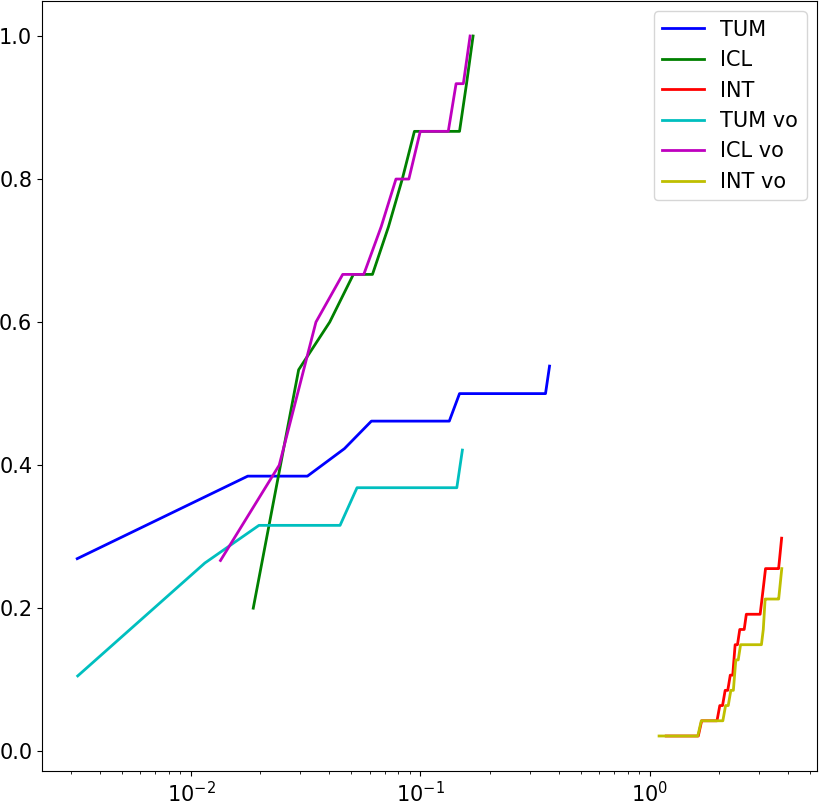}
     \caption{ATE}
     \label{ate}
  \end{subfigure}
  \begin{subfigure}{0.32\textwidth}
    \includegraphics[width=\textwidth]{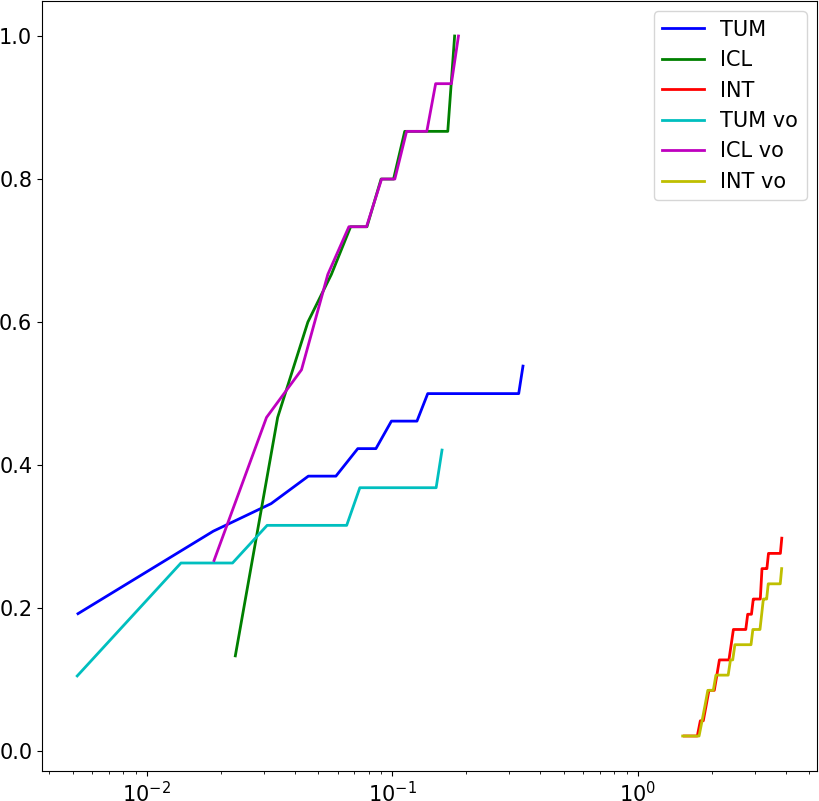}
    \caption{RPE translation}
    \label{rpe-trans}
  \end{subfigure}
  \begin{subfigure}{0.32\textwidth}
    \includegraphics[width=\textwidth]{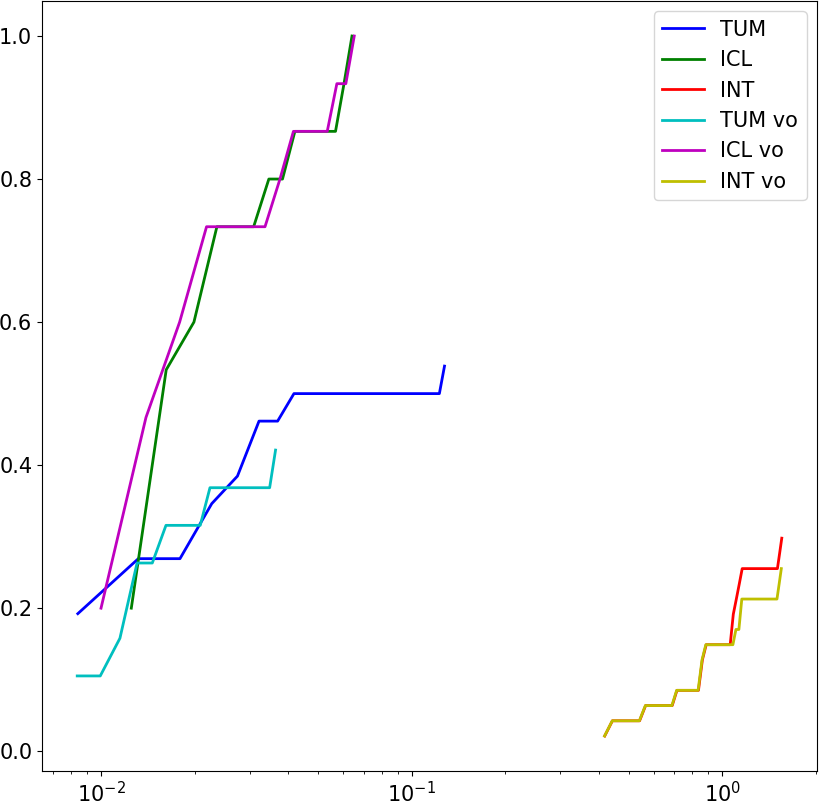}
    \caption{RPE rotation}
    \label{rpe-rot}
  \end{subfigure}
  \caption{We have tested ORBSLAM2 with and without loop closure on 3 open datasets: ICL-NUIM \cite{handa2014benchmark}, TUM RGB-D \cite{sturm2012benchmark} and InteriorNet dataset \cite{li2018interiornet}. Above are bar plots and cumulative plots in logarithmic scale. (a - c) -  bar plots showing the sorted values of the metrics for all trajectories from TUM RGB-D where ORBSLAM2 successfully initialized and did not loose tracking. (d) - Average Trajectory Error (ATE), (e) - Relative Pose Error (RPE) translation, (f) - Relative Pose Error (RPE) rotation. One can see, that the simplest dataset is ICL-NUIM, and the hardest is InteriorNet. Below we perform analysis of the attributes of these datasets to find the reasons of such performance gap.} 
	\label{metrics-all-data}
\end{figure*}
\section{Localization accuracy}
The methods for measuring accuracy of SLAM systems have evolved over a long period \cite{kummerle2009measuring}. For now there is a number of metrics used in research community and sometimes their definitions in papers are misleading. Thus, below we recall the definitions of most popular ATE and RPE metrics.

The motion of a rigid body can be expressed as a sequence in $SE_3$ which gives the transformation from the world to the body frame for each timestamp. $SE_3$ is the group of all rigid transformations in $\mathbb{R}^3$. It is semi-direct product $SO_3 \ltimes \mathbb{R}^3$,  where $SO_3$ is the group of all rotations and $t \in \mathbb{R}^3$ is the 3D Euclidean space. Therefore it has 6 degrees of freedom: 3 for rotation and 3 for translation. The $SE_3$ could be expressed as a pair $(R, t)$ where $R \in SO_3$ and $t \in \mathbb{R}^3$ .
\footnote{More information about $SE_3$ you can find in \cite{eade2013lie} and \cite{blanco2010tutorial}}
Consequently the action of $SE3$ on vectors $x \in \mathbb{R}^3$ is a combination of rotation and translation:
\begin{equation*}
x^{\prime} = Rx + t
\end{equation*}
Homogeneous coordinates are used to provide a matrix representation of the group. So it could be expressed as a $4\times4$ matrix:
\begin{equation*}
M=
\begin{pmatrix}
$R$ & $t$ \\
$0$ & $1$
\end{pmatrix}
\end{equation*}
where $R \in SO_3$ and $t \in \mathbb{R}^3$. It should be noticed that $\forall R \in SO_3 $ is $ R^{-1} = R^T$. Note that orthogonal transformation is isometric. These advices will help you to build an effective algorithm metrics calculation algorithm.

Next, we need three operators to define the metrics: let $M \in SE_3$, then
\begin{equation*}
\mbox{trans}(M) \coloneqq t
\end{equation*}
\begin{equation*}
\mbox{rot}(M) \coloneqq R
\end{equation*}
\begin{equation*}
\angle R \coloneqq \arccos \bigg(\frac{\mathrm{tr}(R)-1}{2}\bigg)
\end{equation*}
For the evaluation, we assume there is a sequence of spatial poses from the estimated trajectory $P_i \in SE_3$ and  from  the  ground  truth  trajectory $Q_i \in SE_3$ in arbitrary world coordinate system. For simplicity, we assume that the both sequences are time-synchronized, and both have length $n$. In practice, these two sequences have different sampling rates, lengths, potentially missing data and in some cases it is necessary to use relative poses, so that an additional data association and pre-processing is required.
\subsection{Absolute trajectory error(ATE)}
The  global  consistency  can  be  evaluated  by  comparing  the absolute  distances  between  estimated  and ground truth trajectory. As  both  trajectories  can  be  specified  in arbitrary  coordinate frames,  they  first  need  to  be  aligned. This  can  be  achieved by using  the Horn method \cite{horn1987closed}, which  finds  the  rigid-body  transformation $S$. Let us define absolute trajectory error matrix at time $i$ as:
\begin{equation*}
E_i \coloneqq Q_i^{-1}SP_i
\end{equation*}
The ATE is defined as the root mean square error from error matrices:
\begin{equation*}
\mbox{ATE}_{rmse} = \bigg(\frac{1}{n} \sum_{i=1}^{n} \parallel \mbox{trans}(E_i) \parallel^2 \bigg)^{\frac{1}{2}}
\end{equation*}
Usually, mean or median values are computed.
Actually, absolute trajectory error is the average deviation from ground truth trajectory per frame.
\subsection{Relative pose error(RPE)}
The relative pose error measures the local accuracy of the trajectory over a fixed time interval $\Delta$. Therefore, the relative pose error corresponds to the drift of the trajectory which is in  particular  useful  for  the  evaluation  of  visual  odometry systems.  Let us define the relative pose error matrix at time step $i$ as:
\begin{equation*}
F_i^{\Delta} \coloneqq (Q_i^{-1} Q_{i+\Delta})^{-1}(P_i^{-1}P_{i+\Delta})
\end{equation*}
from  a  sequence  of $n$ camera  poses we obtain $m = n - \Delta $ individual  relative  pose  error matrices  along  the sequence. The RPE is usually divided into translation and rotation components. Similar  to  the  absolute trajectory  error,  we  propose  to  evaluate the  root  mean  squared  error  over  all  time  indicies for RPE translation error:
\begin{equation*}
\mbox{RPE}_{trans}^{i, \Delta} = \bigg(\frac{1}{m} \sum_{i=1}^{m} \parallel \mbox{trans}(F_i) \parallel^2 \bigg)^{\frac{1}{2}}
\end{equation*}
As for rotation component we use mean error approach:
\begin{equation*}
\mbox{RPE}_{rot}^{i, \Delta} = \frac{1}{m} \sum_{i=1}^{m} \angle (\mbox{rot}(F_i^{\Delta}))
\end{equation*}
For the evaluation of SLAM systems, it makes sense to average over all possible pairs in both translation and rotation component.
\begin{figure*}[t]
  \centering
	\begin{subfigure}{0.32\textwidth}
		\includegraphics[width=\textwidth]{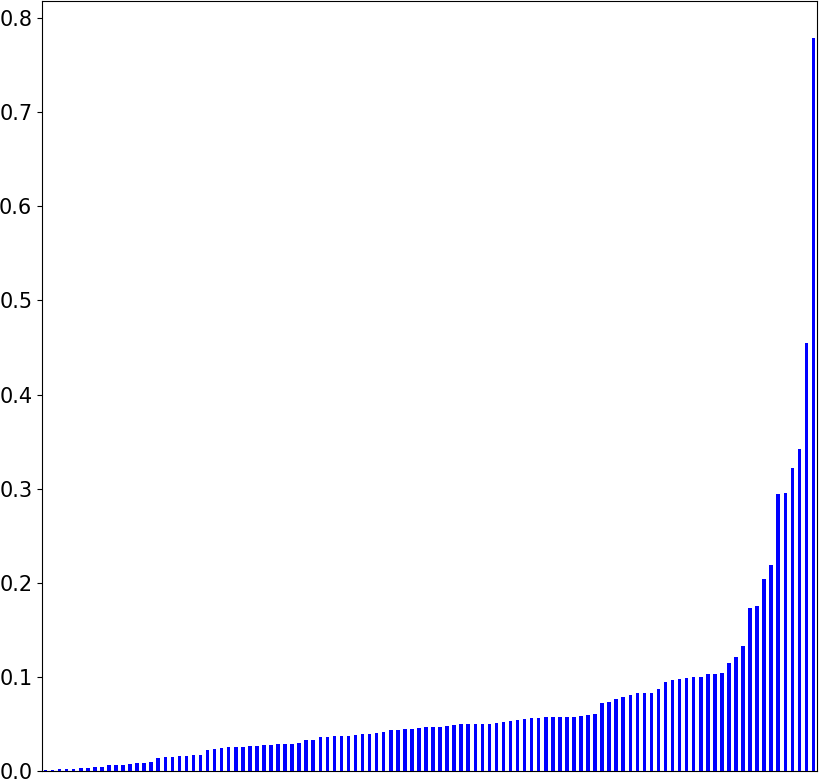}
		\caption{ATE}
		\label{bar-tum-ate}
	\end{subfigure}
	\begin{subfigure}{0.32\textwidth}
    	\includegraphics[width=\textwidth]{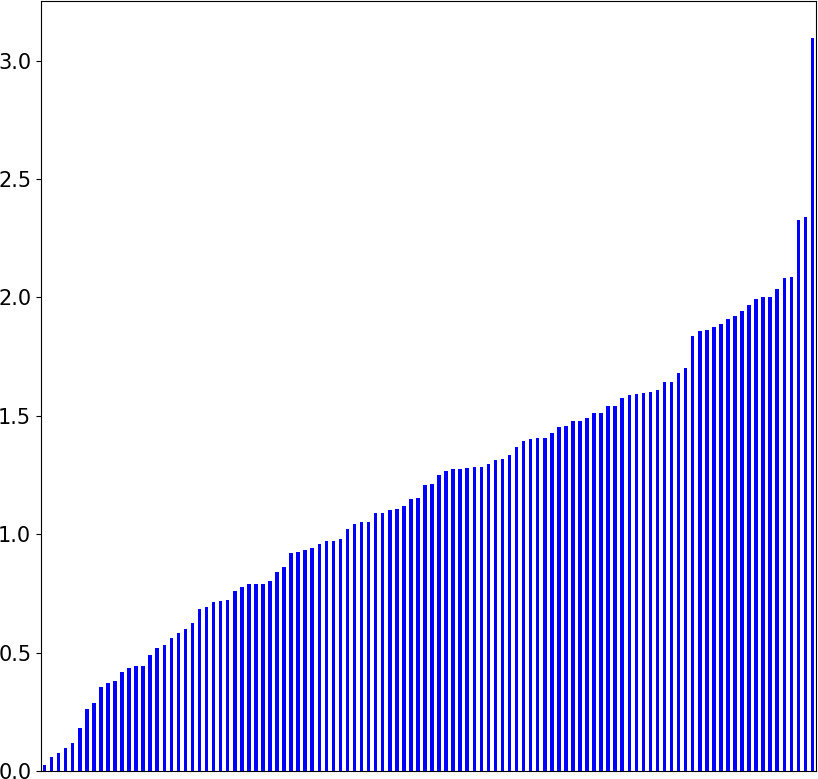}
    	\caption{RPE translation}
    	\label{bar-tum-rpe-trans}
    \end{subfigure}
    \begin{subfigure}{0.32\textwidth}
    	\includegraphics[width=\textwidth]{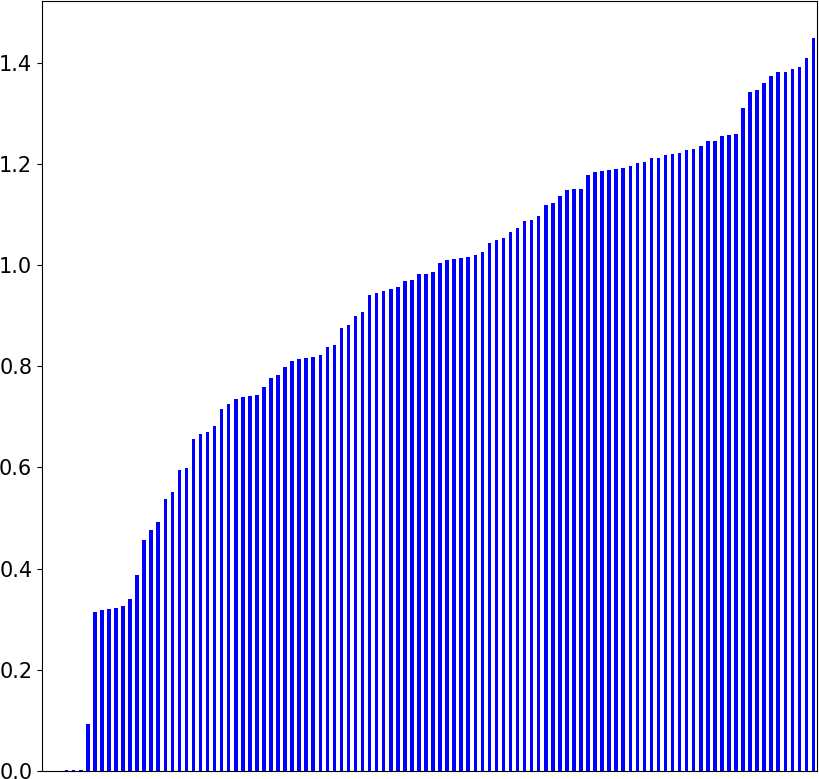}
    	\caption{RPE rotation}
    	\label{bar-tum-rpe-rot}
    \end{subfigure}
	\label{bar-icl-tum}
    \newline
    \centering
    \begin{subfigure}{0.32\textwidth}
        \includegraphics[width=\textwidth]{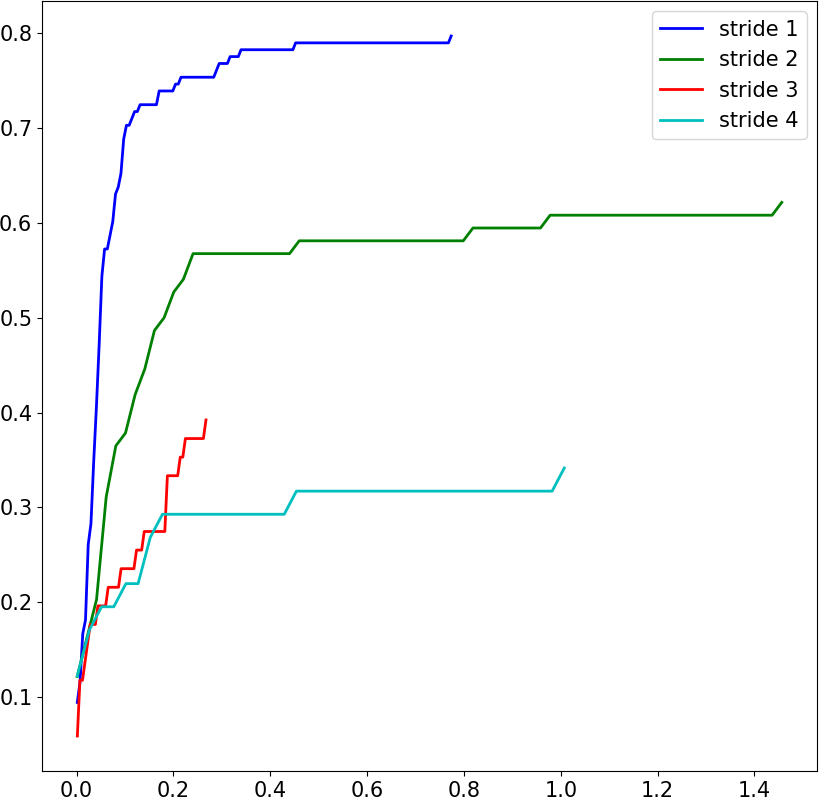}
         \caption{ATE}
      \end{subfigure}
      \begin{subfigure}{0.32\textwidth}
        \includegraphics[width=\textwidth]{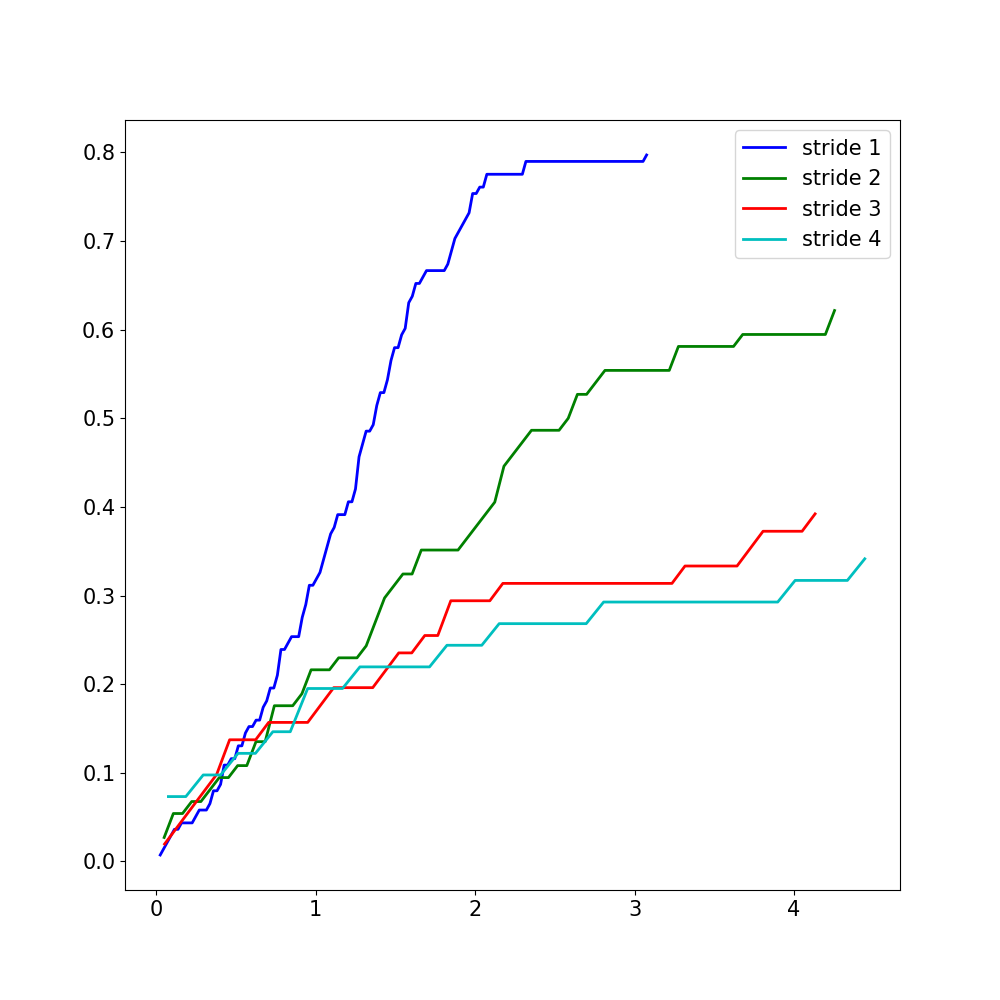}
        \caption{RPE translation}
      \end{subfigure}
      \begin{subfigure}{0.32\textwidth}
        \includegraphics[width=\textwidth]{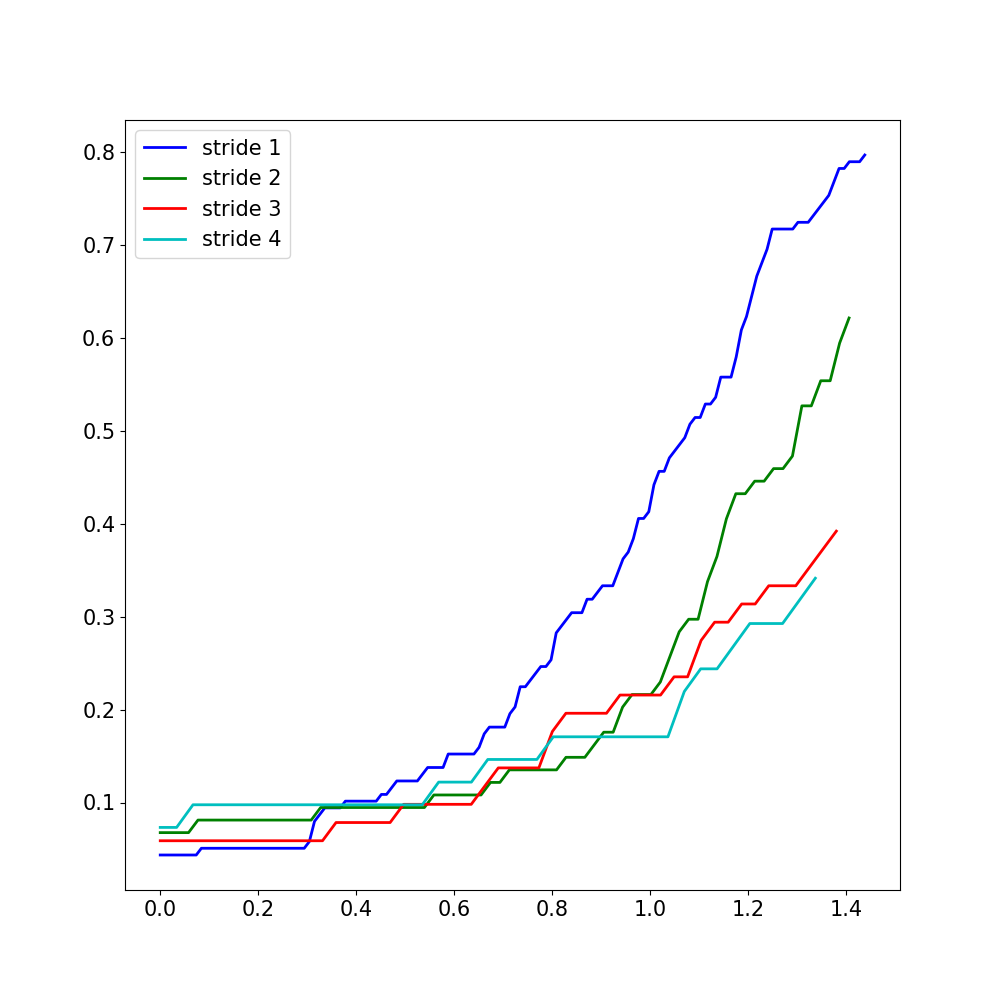}
        \caption{RPE rotation}
      \end{subfigure}
  \caption{Above are the bar plots and cumulative distributions of the metrics for ORBSLAM2 on the new HomeRobot dataset. (a-c) - bar plots show the sorted values of the metrics for all trajectories from HomeRobot dataset where ORBSLAM2 successfully initialized and did not loose tracking. (d) - Average Trajectory Error (ATE), (e) - Relative Pose Error (RPE) translation, (f) - Relative Pose Error (RPE) rotation. We perform experiments on 4 versions of the dataset with different speed of agent's movement. One can see the strong negative correlation between the velocity of the agent and the performance of ORBSLAM2.} 
	\label{metrics-new-data}
\end{figure*}
\begin{figure*}[t]
  \centering
	\begin{subfigure}{0.15\textwidth}
		\includegraphics[width=\textwidth]{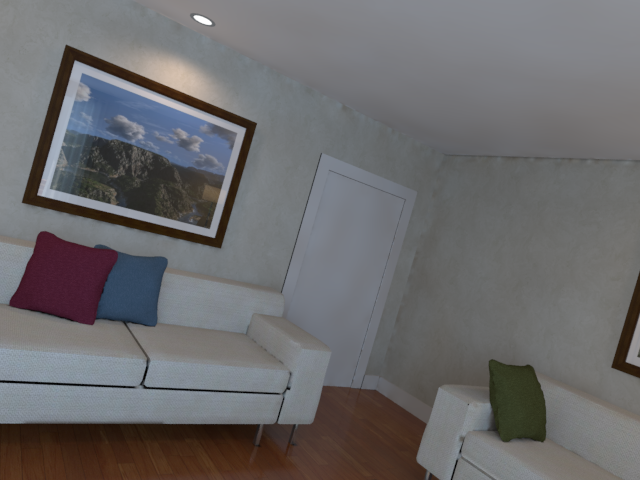}
	\end{subfigure}
	\begin{subfigure}{0.15\textwidth}
		\includegraphics[width=\textwidth]{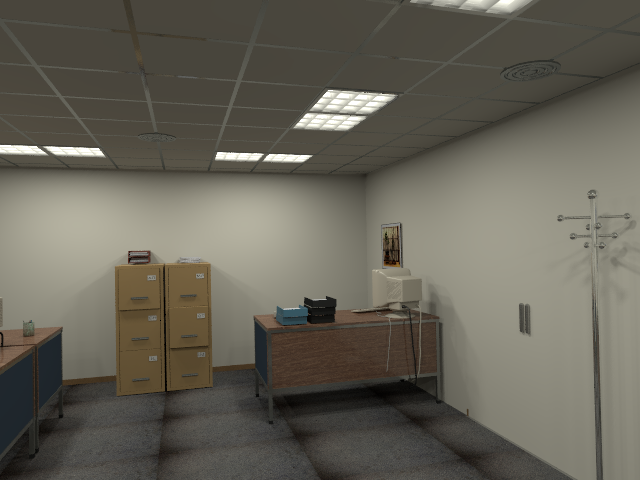}
	\end{subfigure}
	\begin{subfigure}{0.15\textwidth}
		\includegraphics[width=\textwidth]{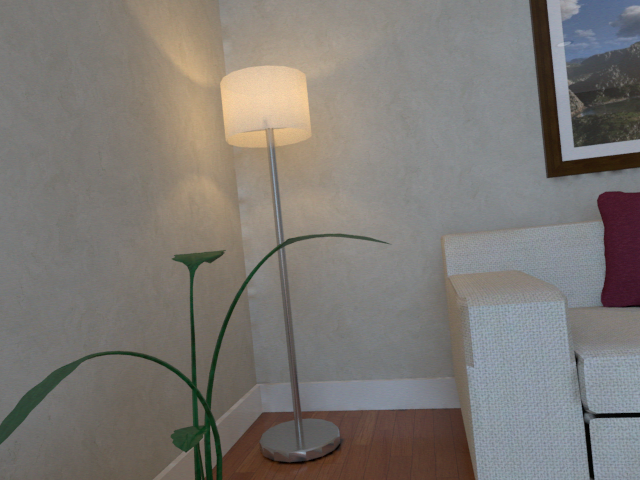}
	\end{subfigure}
	\begin{subfigure}{0.15\textwidth}
    	\includegraphics[width=\textwidth]{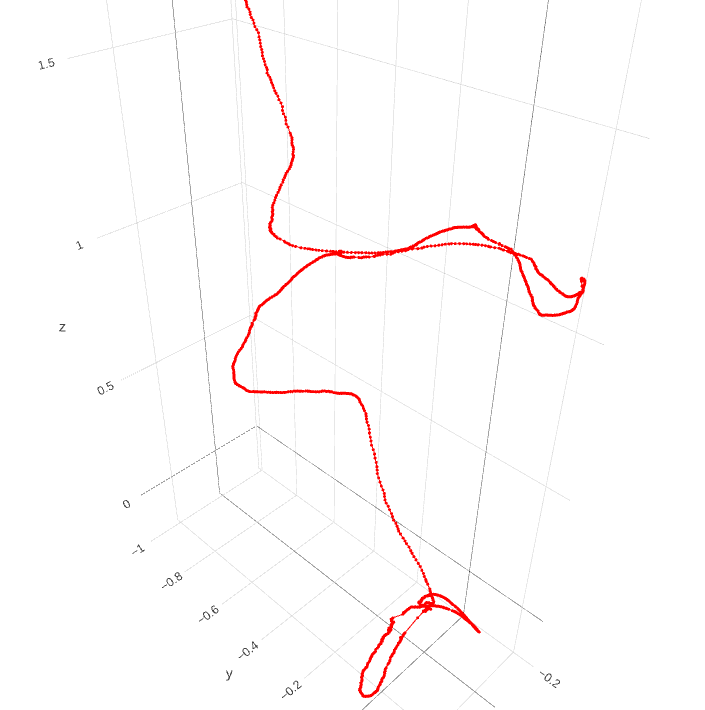}
    \end{subfigure}
	\begin{subfigure}{0.15\textwidth}
    	\includegraphics[width=\textwidth]{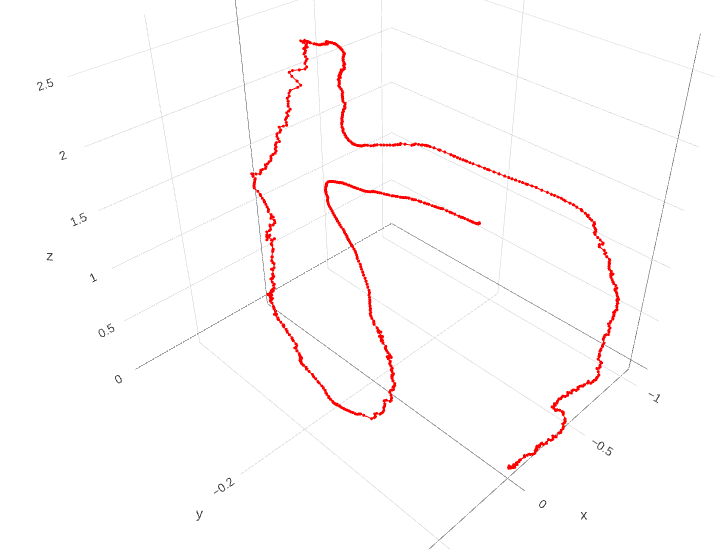}
    \end{subfigure}
	\begin{subfigure}{0.15\textwidth}
    	\includegraphics[width=\textwidth]{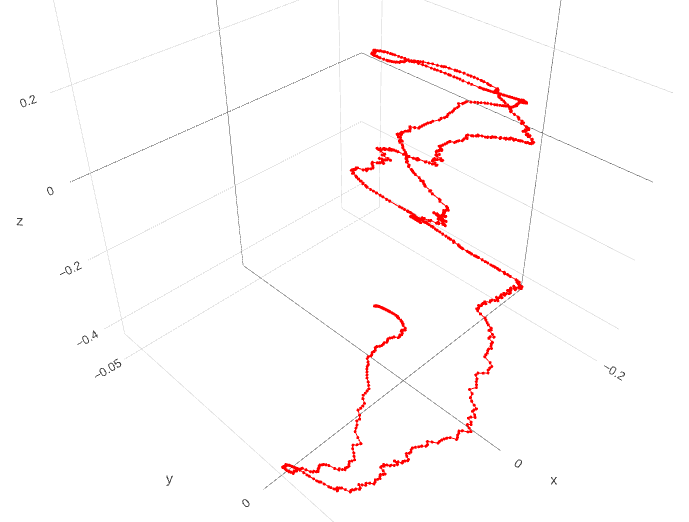}
    \end{subfigure}
    \newline
    \begin{subfigure}{0.15\textwidth}
		\includegraphics[width=\textwidth]{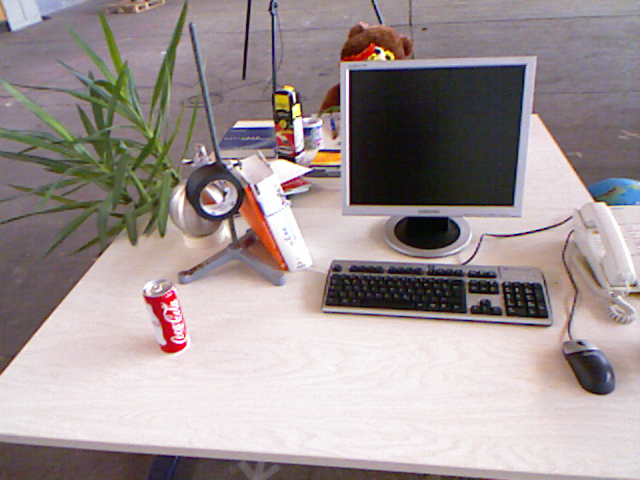}
	\end{subfigure}
    \begin{subfigure}{0.15\textwidth}
		\includegraphics[width=\textwidth]{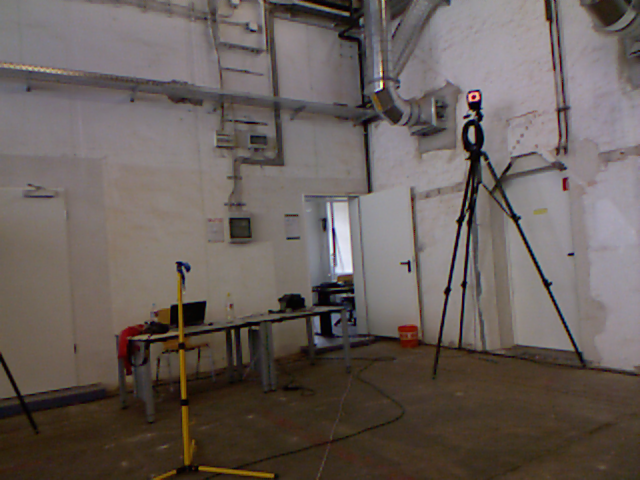}
	\end{subfigure}
    \begin{subfigure}{0.15\textwidth}
		\includegraphics[width=\textwidth]{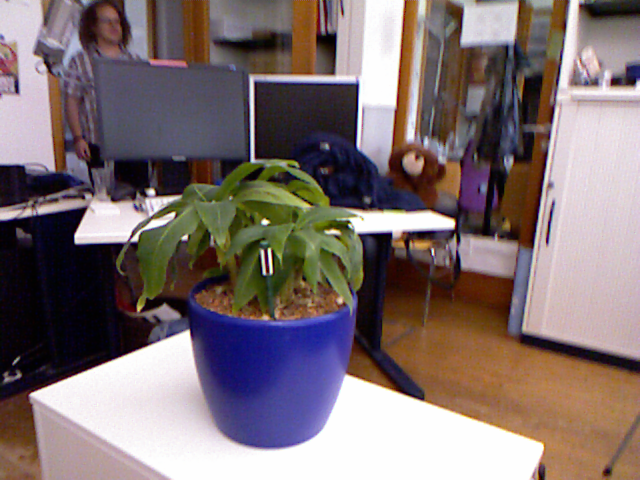}
	\end{subfigure}
	\begin{subfigure}{0.15\textwidth}
    	\includegraphics[width=\textwidth]{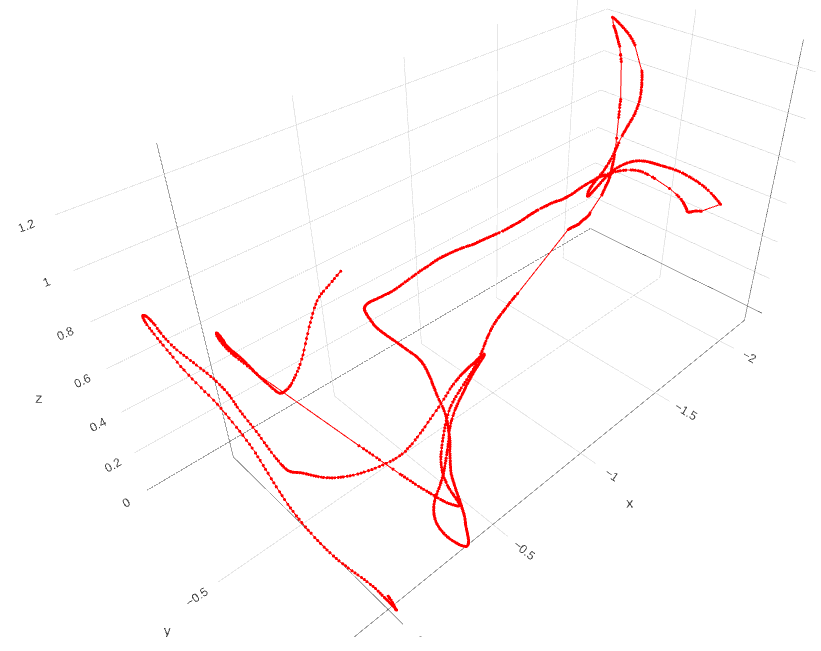}
    \end{subfigure}
	\begin{subfigure}{0.15\textwidth}
    	\includegraphics[width=\textwidth]{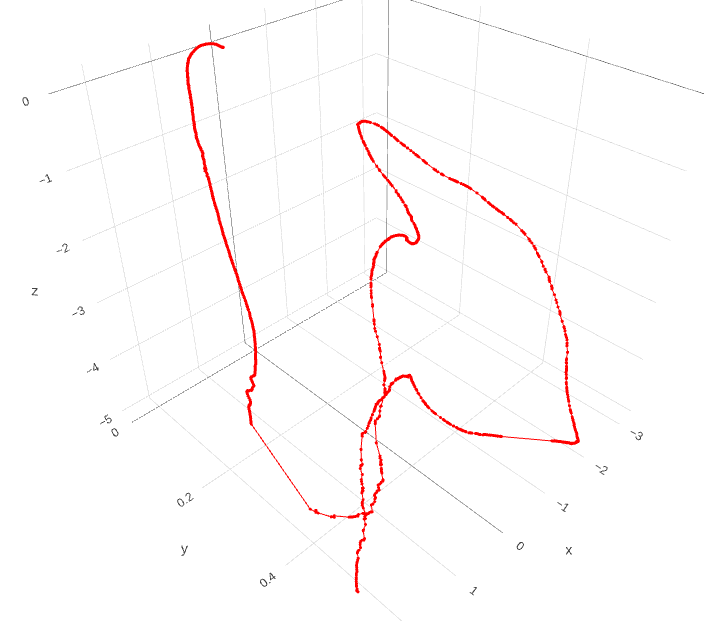}
    \end{subfigure}
	\begin{subfigure}{0.15\textwidth}
    	\includegraphics[width=\textwidth]{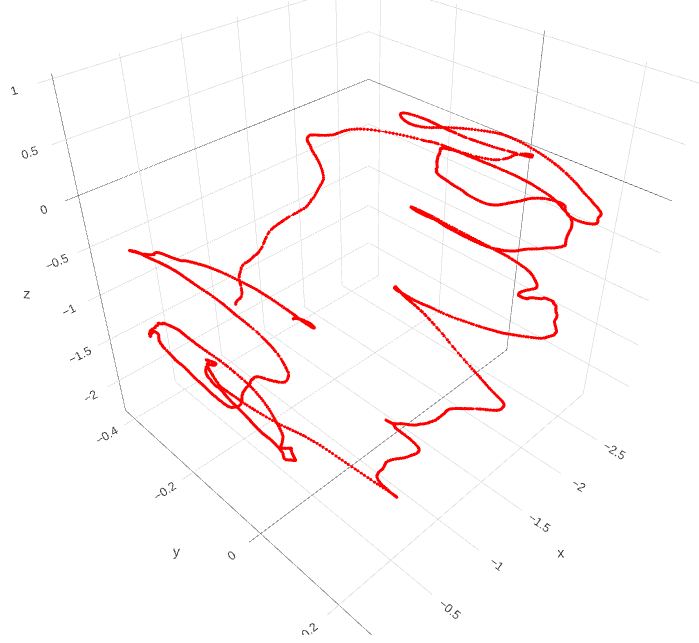}
    \end{subfigure}
    \newline
    \begin{subfigure}{0.15\textwidth}
		\includegraphics[width=\textwidth]{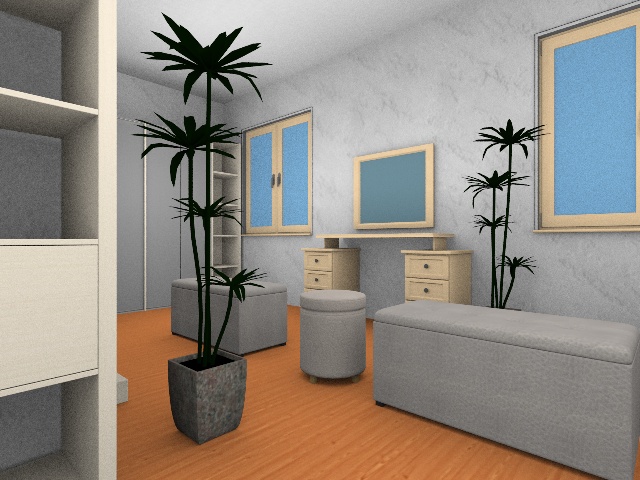}
	\end{subfigure}
    \begin{subfigure}{0.15\textwidth}
		\includegraphics[width=\textwidth]{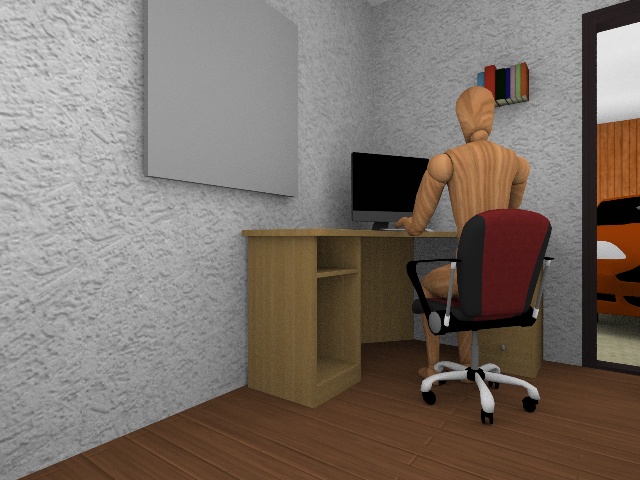}
	\end{subfigure}
    \begin{subfigure}{0.15\textwidth}
		\includegraphics[width=\textwidth]{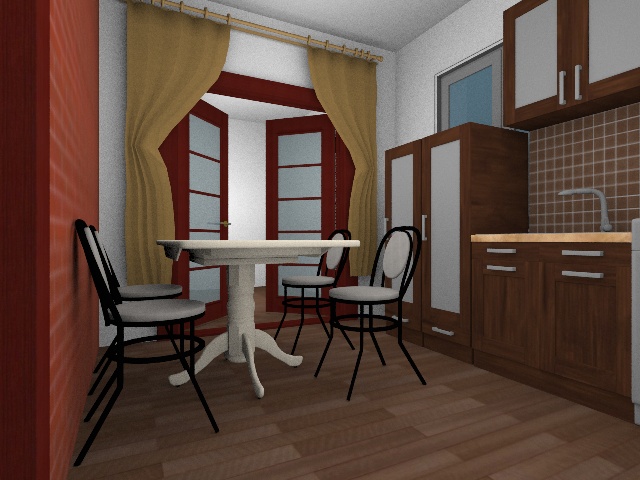}
	\end{subfigure}
	\begin{subfigure}{0.15\textwidth}
    	\includegraphics[width=\textwidth]{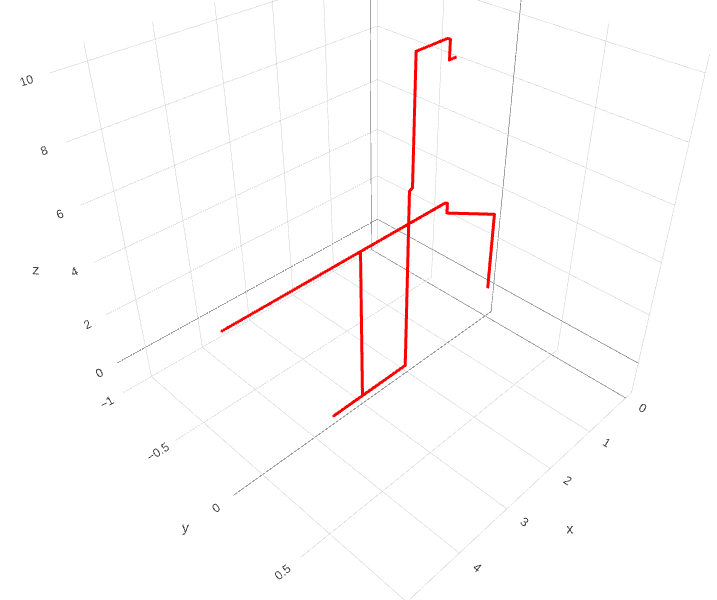}
    \end{subfigure}    
	\begin{subfigure}{0.15\textwidth}
    	\includegraphics[width=\textwidth]{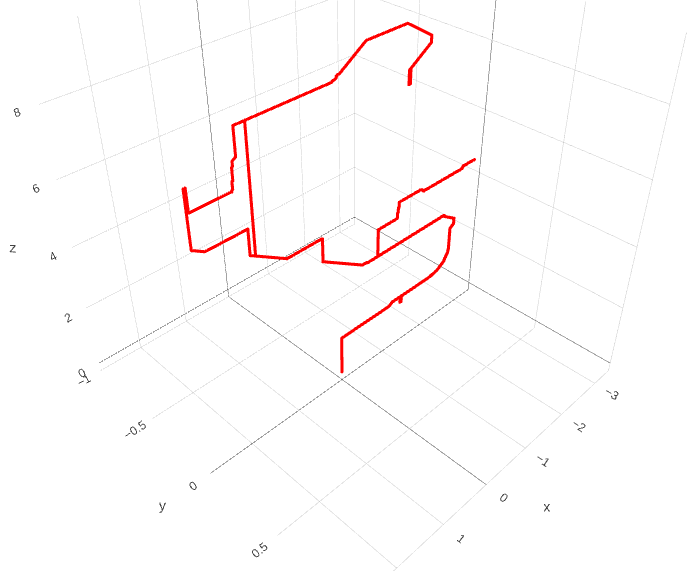}
    \end{subfigure}    
	\begin{subfigure}{0.15\textwidth}
    	\includegraphics[width=\textwidth]{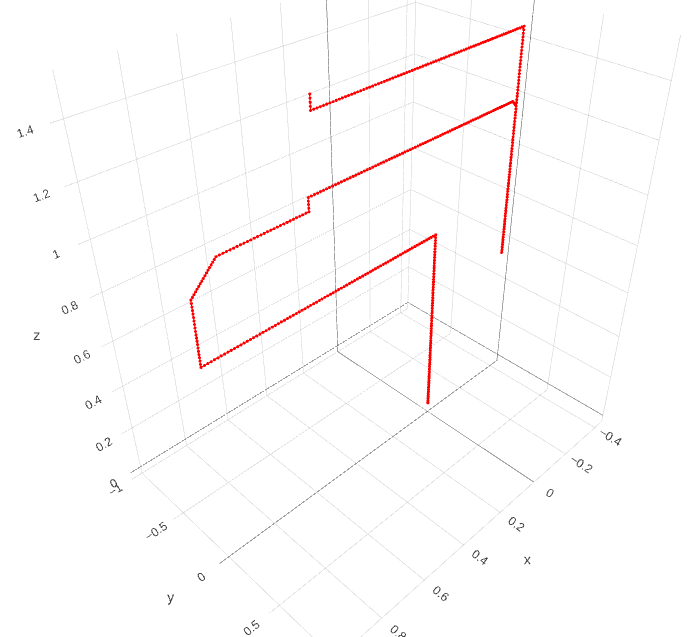}
    \end{subfigure}    
	\caption{Examples of images and trajectories from ICL-NUIM, TUM and HomeRobot datasets.}
	\label{bar-icl-tum}
\end{figure*}
\section{Datasets}
\textbf{TUM RGB-D Benchmark Dataset} \cite{sturm2012benchmark} is a large dataset containing RGB-D data and ground-truth camera poses. TUM dataset contains the RGB and Depth images of Microsoft Kinect sensor along the ground-truth trajectory of the sensor. The data was recorded at full frame rate (30 Hz) and sensor resolution $640\times 480$. The ground-truth trajectory was obtained from a high-accuracy motion-capture system with eight high-speed tracking cameras (100 Hz). TUM dataset contains both a typical office environment (fr1, $6 \times 6m^2$) and a large industrial hall (fr2, $10\times 12m^2$). The benchmark contains 27 sequences.

\textbf{ICL-NUIM RGB-D Benchmark Dataset} \cite{handa2014benchmark} is a synthetic dataset that consists of images obtained from camera in raytraced 3D models in POVRay for two different scenes: the living room and the office room. Images corresponding to different camera poses on a given trajectory are stored in typical RGB-D frame pairs while the camera poses are represented in $SE_3$ format. All data is recorded at 30Hz speed. The benchmark contains 11 sequences.
\begin{table}
  \caption{Characteristics of different datasets. M.vel.p.f - mean velocity per frame; m.ang.v.p.f - mean angular velocity per frame;  m.frames - mean frames.}
  \begin{center}
    \begin{tabular}{c | c c c c}
      \hline
      \hline
      \makebox[8mm]{Dataset} & 
      \makebox[12mm]{m.vel.p.f.} & \makebox[15mm]{m.ang.v.p.f.} &
      \makebox[12mm]{m.frames}\\
      \hline
      TUM  &0.009  & 2.39 & 1424 \\
      ICL-NUIM & 0.007  & 1.29 & 1160 \\
      InteriorNet  & 0.054 & 0.96  & 1000 \\
      HR Ds. skip 0 & 0.006 & 1.45 & 1000\\
      HR Ds. skip 1 & 0.013 & 1.47 & 1000\\
      HR Ds. skip 2 & 0.019 & 1.49 & 1000\\
      HR Ds. skip 3 & 0.025 & 1.51 & 1000\\
      \hline
      \hline
    \end{tabular}
    \label{orb-table}
  \end{center}
\end{table}

\textbf{InteriorNet  Photo-realistic Indoor Scenes Dataset} \cite{li2018interiornet} is an end-to-end pipeline to render photo-realistic images and associated ground truth data. Were created multiple trajectories for each layout by varying the combination of linear velocity, angular velocity and trajectory types. Images support 640x480 resolution, at 25 Hz, resulting in 1,000 images per trajectory. Depth is obtained as the Euclidean distance of the ray intersection; and noisy depth is generated by simulating a real Kinect mechanism. The publicly available part of the dataset contains 40 sequences.

\textbf{HomeRobot Dataset}
is a novel dataset that we have generated for the experiments. It contains long sequences sampled using a large 3D model repository for indoor scenes SUNCG \cite{song2016ssc}. We have generated randomized trajectories of the moving agent using the occupancy grids provided by the 3D models. The trajectories emulate the movements of a home robot with a camera mounted at a fixed height above the floor. Since the probability of failure of ORBSLAM2 is higher for long sequences, we have sliced the trajectories into overlapping chunks of fixed length (we have chosen 1000 frames as a chunk length similary to the other RGB-D datasets) in order to enable more fair evaluation. We used our custom rendering engine with ambient occlusion technique that allowed us to get high quality images. The effects of the noise in depth estimation are modelled similarly to the ICL-NUIM. We have generated the sequences of HomeRobot dataset such that the speed of the agent is approximately the same as in TUM RGB-D. Overall, the new benchmark contains around 1000 sequences.
\section{Experiments and analysis}
\begin{table}[t]
  \caption{Comparison of our results with original results from \cite{mur2017orb} in terms of ATE.}
  \begin{center}
    \begin{tabular}{c | c c}
      \hline
      \hline
      \makebox[20mm]{Traj. name} & \makebox[15mm]{Our ATE} & \makebox[20mm]{Results from \cite{mur2017orb}} \\
      \hline
      TUM fr1 xyz & 0.010m & 0.010m \\
      TUM fr1 desk & 0.016m & 0.016m \\
      TUM fr2 xyz & 0.004m & 0.004m \\
      ICL of. tr1 & 0.053m & 0.051m \\
      \hline
      \hline
    \end{tabular}
    \label{sample-table}
  \end{center}
\end{table}

\textbf{Experiments on the open benchmarks.}
First, we proved our metrics measuring methods, we compared our results on popular trajectories with results from \cite{mur2017orb}. The results of the comparison are shown in the Table \ref{sample-table}. Second, we perform extensive evaluation of ORBSLAM2 on the open benchmarks. We use the following evaluation protocol. Since ORBSLAM2 is randomized, each sequence in the dataset is run 10 times to find the median value for each metric. We build cumulative plots, which show the total number of trajectories, expressed in percentage, that have the metric value less than or equal to a series of thresholds.

The experiments have shown that there is a threshold value of the metrics indicating that the algorithm has failed on trajectory. There is a noticeable gap between the values of the metrics Figure \ref{metrics-all-data}. We have visualized and inspected the estimated trajectories and confirmed the existence of complexity gap. So there is a threshold value of metrics, corresponding to a dataset, from which ORBSLAM2 did not work properly. We assume that the threshold value depends on the dataset parameters (Table \ref{orb-table}), such as velocity and angular velocity of agent. This assumption is confirmed by our degrading experiment on our new dataset Figure \ref{metrics-new-data}.

\textbf{Cumulative distribution analysis.}
 Figure \ref{metrics-all-data} shows the cumulative plots for ICL-NUIM, TUM RGB-D, and InteriorNet datasets. Easy to see that
ICL-NUIM is simple dataset for ORBSLAM2, as InteriorNet is the hardest. The loop detection and relocalization did not give significant gain in accuracy Figure \ref{metrics-all-data}. The effect is visible only in hard cases, where method works badly with or without loop closure.

\textbf{Experiments of the HomeRobot dataset}
We have used the same protocol for evaluation of ORBSLAM2 on the new HomeRobot dataset. We varied the speed of the agent by skipping frames in the sequences and built separate cumulative plots for each value of the stride parameter. Figure \ref{metrics-new-data} shows the resulting cumulative plots for ATE, RPE rotation and RPE translation. Thus higher velocity of the agent results in lower accuracy of ORBSLAM2, due to reduction of the overlap between the consecutive frames.
\section{Discussion and future work}
In this paper we analyzed robustness of modern RGB-D SLAM methods on a large scale. We analyzed cumulative distributions of ATE and RPE metrics for renowned ORBSLAM2 system across popular open benchmarks including more than a hundred trajectories in total. Our experiments show that while accuracy on some sequences is high, the robustness of SLAM to changing conditions is an overlooked issue.

We also introduce a new HomeRobot dataset that is designed for indoor robot navigation tasks. The dataset and the source code for reproducing our experiments will be publicly available at the time of publication. We also plan to generate larger datasets of synthetic data for training and validation of the learning-based visual SLAM methods. We believe our results would be of value as a framework for future research and developing new methods for visual SLAM.


\bibliographystyle{plain}
\bibliography{references}

\end{document}